\documentclass[10pt,twocolumn,letterpaper]{article}

\usepackage{stfloats}

\usepackage[pagenumbers]{wacv} % To force page numbers, e.g. for an arXiv version

\usepackage{booktabs}
\usepackage{multirow}
\usepackage{float} 
\definecolor{wacvblue}{rgb}{0.21,0.49,0.74}
\usepackage[pagebackref,breaklinks,colorlinks,allcolors=wacvblue]{hyperref}

\def\wacvPaperID{2241} % *** Enter the WACV Paper ID here
\def\confName{WACV}
\def\confYear{2026}

\title{VividAnimator: An End-to-End Audio and Pose-driven Half-Body Human Animation Framework
}

\author{
Donglin Huang$^{1}$\textsuperscript{*},  Yongyuan Li$^{2,\dag}$\textsuperscript{*}, Tianhang Liu$^1$\textsuperscript{*} \\
Junming Huang$^1$, Xiaoda Yang$^{1}$, Chi Wang$^1$, Weiwei Xu$^{1,\ddag}$ \\
$^1$Zhejiang University, $^2$Image Derivative Inc \\
{\tt\small huangdonglin@zju.edu.cn,} {\tt\small liyongyuan@idr.ai,} {\tt\small tianhang@zju.edu.cn}  \\
{\tt\small \{junminghuang, xiaodayang, chi.wang\}@zju.edu.cn,}
{\tt\small xww@cad.zju.edu.cn}
}

\begin{document}

\twocolumn[{%

\maketitle

\begin{center}
    \centering
    \captionsetup{type=figure}
    \includegraphics[width=\textwidth,trim=0cm 0cm 0cm 0cm, clip]{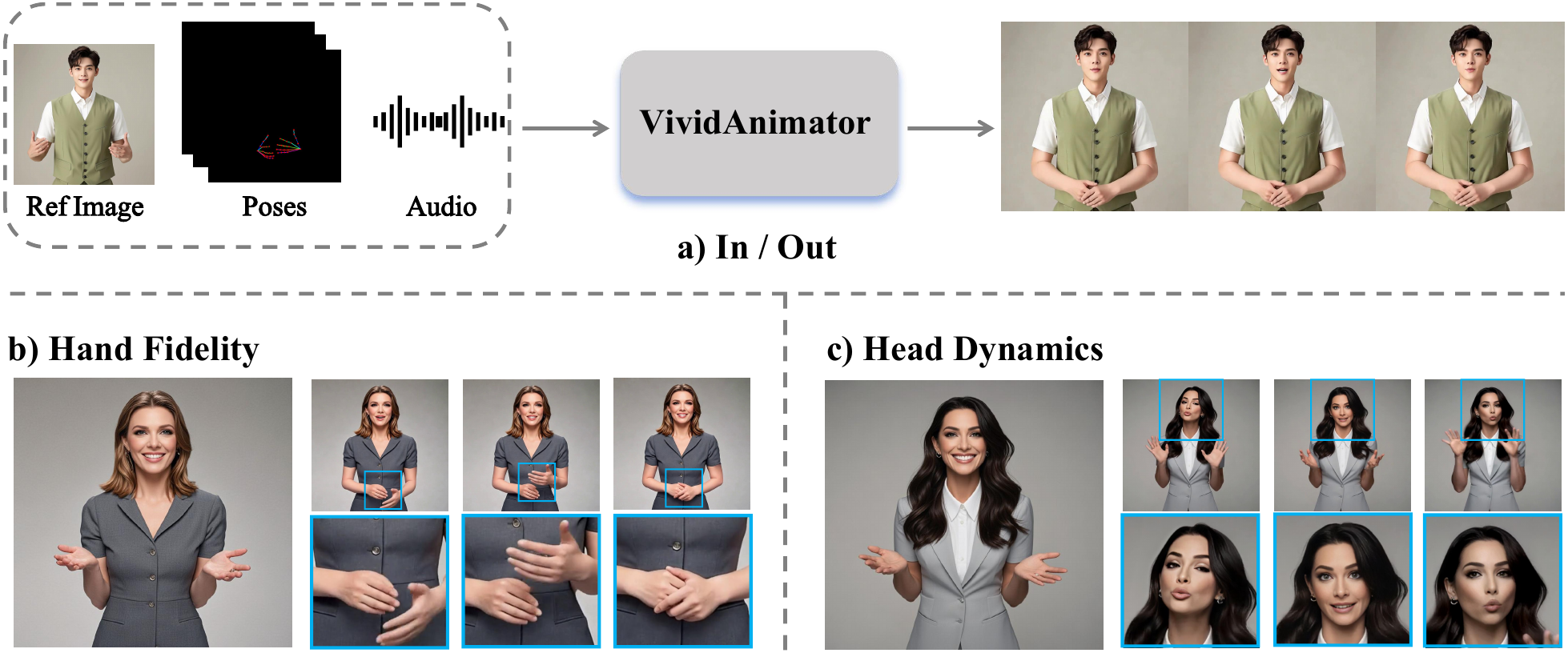}
    \vspace{-1.0em}
    \captionof{figure}{
        a) Vivid Animator synthesizes high-fidelity, identity-preserving videos with rich and expressive motion. Given a reference image, a sequence of poses, and audio, our method showcases its capability to produce: b) precise hand reconstruction and c) dynamic, natural head motion synchronized with speech.
    }
    \label{fig:1}
\end{center}
}]%
\makeatletter\def\Hy@Warning#1{}\makeatother
{\let\thefootnote\relax\footnotetext{\noindent\textsuperscript{*} Equal contribution.}
{\let\thefootnote\relax\footnotetext{\noindent\dag\ Project lead.}
{\let\thefootnote\relax\footnotetext{\noindent\ddag\ Corresponding author.}

\begin{abstract}
Existing for audio- and pose-driven human animation methods often struggle with stiff head movements and blurry hands, primarily due to the weak correlation between audio and head movements and the structural complexity of hands.
To address these issues, we propose \textbf{VividAnimator}, an end-to-end framework for generating high-quality, half-body human animations driven by audio and sparse hand pose conditions. Our framework introduces three key innovations. First, to overcome the instability and high cost of online codebook training, we pre-train a Hand Clarity Codebook (HCC) that encodes rich, high-fidelity hand texture priors, significantly mitigating hand degradation. Second, we design a Dual-Stream Audio-Aware Module (DSAA) to model lip synchronization and natural head pose dynamics separately while enabling interaction. Third, we introduce a Pose Calibration Trick (PCT) that refines and aligns pose conditions by relaxing rigid constraints, ensuring smooth and natural gesture transitions. Extensive experiments demonstrate that Vivid Animator achieves state-of-the-art performance, producing videos with superior hand detail, gesture realism, and identity consistency, validated by both quantitative metrics and qualitative evaluations.
\end{abstract}
    
\section{Introduction}
\label{sec:intro}
The task of human animation specifically focuses on generating a realistic video of a person from a single reference image, leveraging driving signals like audio, text, or a target video. Unlike talking head generation~\cite{loopy,sonic,hdtr,hallo,echomimic}, which primarily focuses on animating the facial region, human animation need synthesize complex full-body movements, including the articulation of limbs and hands. This capability unlocks applications in entertainment, film production, and virtual character creation. However, espite recent progress enabled by diffusion-based generative models, current methods still struggle with challenges.  For example, the generated videos often lack fine-grained texture details and motion rhythm, thereby limiting the widespread application of these approaches.
       
In recent years, human animation has been extensively studied, with pose-transfer methods~\cite{hu2024animateanyoneconsistentcontrollable,tu2024stableanimatorhighqualityidentitypreservinghuman,zhang2025mimicmotionhighqualityhumanmotion,wang2024unianimate,shao2024human4dit,chang2024magicposerealistichumanposes} emerging as a dominant approach due to their controllability and structural consistency. By conditioning on a predefined sequence of poses, these methods can effectively guide human video generation.
However, pose-transfer methods still face two critical limitations. First, generated hand regions often appear blurry and lack fine-grained details.
Second, their rigid reliance on a full pose skeleton restricts their ability to produce spontaneous, natural gestures that could be better inferred from other signals, such as audio, which ultimately limits the expressiveness of the animation.
Beyond pose-driven methods, researchers have also explored multi-modal conditions such as audio and text~\cite{cui2025hallo3highlydynamicrealistic, meng2025echomimicv3,gan2025omniavatarefficientaudiodrivenavatar, kong2025lettalkaudiodrivenmultiperson}. While these conditions have been explored for their potential to drive human motion synthesis, their application is often hindered by the weak correlation between the input signal and body movements. As a result, current models tend to learn an averaged data distribution, leading to a significant lack of diversity and unpredictability of motion in the generated videos.

In this paper, we introduce Vivid Animator, an end-to-end framework for high-fidelity half-body human animation. Our method aims to simultaneously improve rendering quality and generate expressive, natural movements. To achieve this, Vivid Animator integrates three key components: 1) a pre-trained VAE model, dubbed the Hand Clarity Codebook (HCC), to enhance hand texture details by injecting features that incorporate rich hand priors; 2) a Dual-Stream Audio-Aware Module (DSAA) that leverages cross-attention to fuse audio and motion features, thereby enhancing the expressiveness of rhythmic head movements; 3) a Pose Calibration Trick (PCT), which achieves precise pixel-space alignment by applying geometric fitting and scale calibration. Each of these components is detailed below.

First, we employ a pre-trained VAE model, referred to as the Hand Clarity Codebook (HCC) that has been trained on a large dataset of high-quality hand images. Unlike CyberHost~\cite{lin2025cyberhost}, a key limitation of online codebook training is the mutual influence on model convergence, which can lead to training instability. Furthermore, the need to synchronously optimize both networks significantly increases computational complexity. 
The Hand Clarity Codebook significantly enhances hand texture details by encoding rich hand prior knowledge. Specifically, given cropped hands from the reference image, we perform a nearest-neighbor search within the codebook to retrieve a discrete vector. This vector is then injected into denoising UNet to enhance the fine-grained texture details of the hands, as shown in Fig.~\ref{fig:1} b).

Secondly, to address the issue of stiff head poses and generic gestures caused by weak audio-motion correlation, we propose a Dual-Stream Audio-Aware module. The module processes audio through two specialized branches, one focusing on lip synchronization and another for modeling head motion. This allows our model to produce more expressive and lifelike animations, with head movements that naturally synchronize with the rhythmic patterns of speech, as shown in Fig.~\ref{fig:1}c). 

Additionally, we propose a Pose Calibration Trick (PCT) to effectively resolve pose misalignment in pose-driven methods that may result in artifacts. The PCT improves motion coherence by applying geometric fitting and scale calibration to align the driving pose with the reference image.

Our main contributions are summarized as follows:
\begin{itemize}
    \item We introduce a end-to-end diffusion framework, Vivid Animator, for high-fidelity half-body human animation that simultaneously ensures superior rendering quality and expressive motion generation.
    \item We propose a Hand Clarity Codebook (HCC), a VAE-based model pre-trained on a large, high-quality dataset, which significantly enhances fine-grained hand texture details by injecting rich hand priors into the denoising UNet.
    \item We design a Dual-Stream Audio-Aware Module (DSAA) that decouples audio processing into two specialized branches, enabling the generation of more natural and expressive head movements that are aligned with speech prosody.
    \item Extensive experiments and analyses demonstrate the effectiveness of our method, surpassing the current state-of-the-art approaches.
\end{itemize}
% \secvspace
\section{Related Work}
% \secvspace
\label{sec:related_work}

\subsection{Pose-Driven Human Animation}
Recent studies~\cite{hu2024animateanyoneconsistentcontrollable,chang2024magicposerealistichumanposes,xu2023magicanimatetemporallyconsistenthuman} have adopted a dual-network design, where a pose network encodes skeletal motions~\cite{yang2023effective,denseposedensehumanpose} to guide animation, while a reference network preserves identity information. Beyond skeletal poses, other approaches~\cite{zhu2024champcontrollableconsistenthuman,shao2024human4dit,yu2024signavatarslargescale3dsign,shottalkwholebodytalking} leverage additional driving signals such as depth maps and 3D parametric models~\cite{Bogo:ECCV:2016,pavlakos2019expressivebodycapture3d,cai2024smplerxscalingexpressivehuman}, which provide explicit geometric priors that enhance alignment and improve motion guidance. For instance, RealisDance~\cite{zhou2024realisdanceequipcontrollablecharacter} introduces a pose-gating module conditioned on multiple pose types, including HaMeR~\cite{hamer} and SMPL-X~\cite{cai2024smplerxscalingexpressivehuman}, thereby enriching geometric information. 
TALK-Act~\cite{guan2024talkactenhancetexturalawareness2d} integrates Hand-Attention layers into the U-Net to improve the preservation of hand structure. To further enhance stability and identity fidelity, MimicMotion~\cite{zhang2025mimicmotionhighqualityhumanmotion} employs confidence-aware pose guidance together with a regional loss amplification strategy to mitigate image distortion. StableAnimator~\cite{tu2024stableanimatorhighqualityidentitypreservinghuman} adopts a cross-attention mechanism to inject ArcFace features~\cite{ARCFACE} with a face mask, thereby preserving facial structure and identity. 
HandRefiner~\cite{lu2024handrefinerrefiningmalformedhands} exploits 3D hand mesh reconstruction~\cite{lin2021meshgraphormer} as a control signal, integrated into the U-Net through ControlNet~\cite{zhang2023adding}, to achieve more precise hand generation. ShowMaker~\cite{yang2024showmaker} further improves fidelity by introducing a keypoint-based hand modeling module that captures fine-grained hand motion and a face recapture module that maintains identity consistency.

\subsection{Audio Driven Human Animation}
Audio driven human animation~\cite{liang2024superior, tian2025emo2endeffectorguidedaudiodriven,wang2025fantasytalking2timesteplayeradaptivepreference, qi2025chatanyonestylizedrealtimeportrait, loopy, gan2025omniavatarefficientaudiodrivenavatar,wei2024aniportrait,kong2025lettalkaudiodrivenmultiperson} aims to synthesize facial expressions, lip movements, and body gestures that match the semantics, emotion, and rhythm of an input audio signal. 
Research in this domain mainly falls into two categories: \textit{talking head animation} and \textit{co-speech full-body animation}.
In the field of talking head animation, diffusion models have become the dominant method. 
Vlogger~\cite{zhuang2024vlogger} explores generating animated portraits from a single static image.
EMO~\cite{emo} enhances temporal coherence using a frame encoding module.  
MegActor-$\Sigma$~\cite{yang2024megactorsigmaunlockingflexiblemixedmodal} integrates audio and visual features into a conditional diffusion Transformer for more realistic lip and expression synthesis. 
Hallo3~\cite{cui2025hallo3highlydynamicrealistic} builds upon a pretrained Transformer, incorporating a causal 3D VAE and an identity reference network to improve expression dynamics and identity preservation. 
These methods have significantly improved lip-sync accuracy and the realism of facial expressions.
While facial animation focuses on a limited region, co-speech full-body animation presents a greater challenge, as it requires the synthesis of dynamic, full-body motion.
TANGO~\cite{liu2024tangocospeechgesturevideo} focuses on learning an audio-to-gesture mapping while minimizing visual artifacts.
By utilizing a sample hand pose as guidance, EchoMimicV2~\cite{meng2025echomimicv2strikingsimplifiedsemibody} achieves half-body animation.
CyberHost~\cite{lin2025cyberhost} supports a broader range of control signals, including audio, full-body keypoints, and hand poses. 
OmniHuman~\cite{lin2025omnihuman1rethinkingscalinguponestage} employs a multi-condition control framework to ensure high consistency and controllability outputs.
\section{Methods}
\label{sec:method}

\begin{figure*}[t]
    \centering
    \includegraphics[width=0.86\linewidth,trim=0.7cm 0.5cm 0.5cm 0cm, clip]{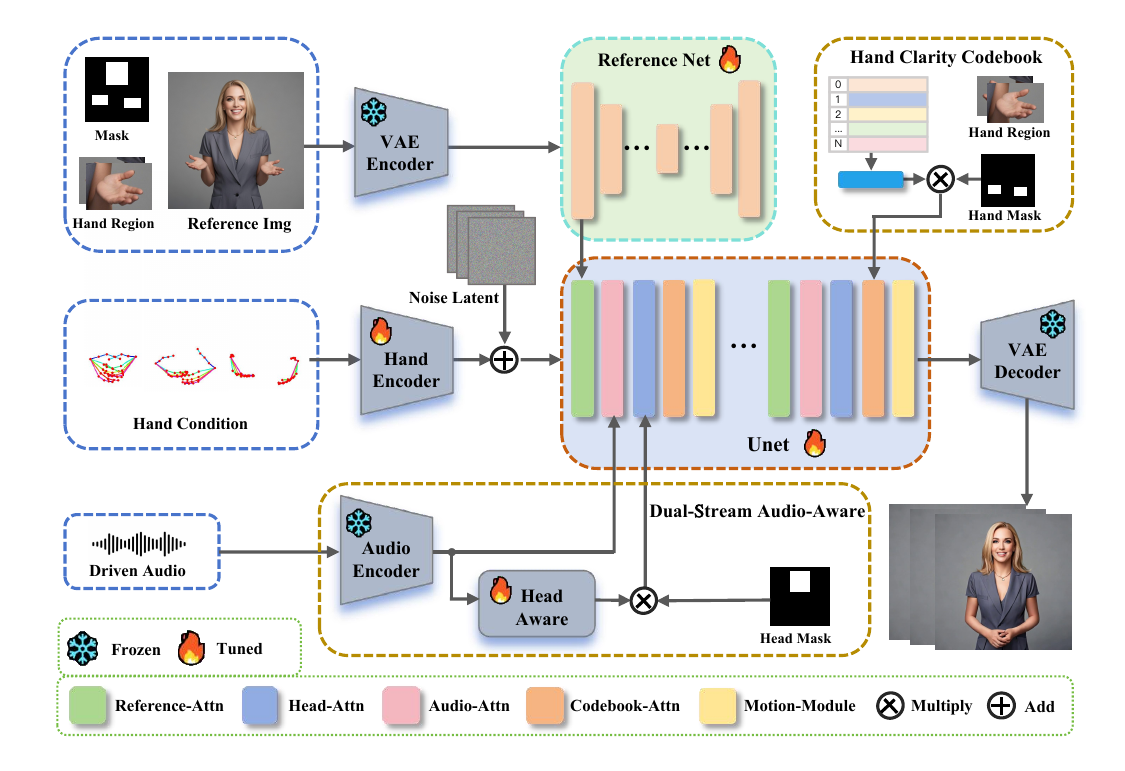}
    \caption{
    Overview of Vivid Animator, a pipeline for realistic half-body human animation.
    Given a reference image, an audio clip, and a hand pose sequence as input, our method synthesizes high-fidelity videos with natural motion. A Hand Clarity Codebook (HCC) is introduced to preserve fine-grained hand textures, while a Dual-Stream Audio-Aware Module (DSAA) enables rhythmic and coherent head motion.}
    \label{fig:2}
\end{figure*}

%-------------------------------------------------------
\subsection{Preliminaries}

\paragraph{Latent Diffusion Model.}
We build our framework upon the LDM~\cite{rombach2022highresolutionimagesynthesislatent}, which operates in the latent space of a pretrained Variational Autoencoder (VAE)~\cite{kingma2022autoencodingvariationalbayes}. 
Given an input image $I$, the VAE encoder $\mathcal{E}$ maps it into a compact latent representation $z_{0} = \mathcal{E}(I)$, which significantly reduces the computational cost compared to diffusion in pixel space.  

During training, Gaussian noise is gradually added to the latent variable across timesteps $t \in \{1, \ldots, T\}$. The forward process is defined as:  

\begin{equation}
q(\mathbf{z}_{t} \mid \mathbf{z}_{t-1}) 
= \mathcal{N}\!\left(\sqrt{1-\beta_t}\,\mathbf{z}_{t-1},\, \beta_t \mathbf{I}\right)
\label{eq:ldm_forward}
\end{equation}

where $\beta_t$ denotes the variance schedule.  
The reverse process is parameterized by a conditional denoising network $\Phi_{\theta}$, which predicts the clean latent from the noisy one:  

\begin{equation}
\mathbf{z}_{t-1} = \Phi_{\theta}(\mathbf{z}_{t},\, t,\, \mathbf{c})
\label{eq:ldm_reverse}
\end{equation}

with $\mathbf{c}$ representing external conditions such as audio, reference images, or pose keypoints.  
The training objective is to minimize the discrepancy between the injected noise $\epsilon$ and the network’s prediction at every timestep:  

\begin{equation}
L = \mathbb{E}_{\mathbf{z}_t,\,t,\,\mathbf{c},\,\epsilon \sim \mathcal{N}(0,1)} 
\Big[ \| \epsilon - \epsilon_{\theta}(\mathbf{z}_t, t, \mathbf{c}) \|^2_2 \Big]
\label{eq:ldm_loss}
\end{equation}

where $\epsilon_{\theta}$ denotes the denoising U-Net with trainable parameters $\theta$.  
At inference time, the model progressively removes noise from a Gaussian sample $\mathbf{z}_T$, and the VAE decoder $\mathcal{D}$ reconstructs the final image or video frames from the denoised latent $\mathbf{z}_0$.  

% \paragraph{General Pose-Driven Framework.}
\subsection{Network Architecture}

\paragraph{Method Overview.}
Fig.\ref{fig:2} provides an overview of our method. Following recent works on pose-driven generation~\cite{chang2024magicposerealistichumanposes,xu2023magicanimatetemporallyconsistenthuman,zhou2024realisdanceequipcontrollablecharacter,hu2024animateanyoneconsistentcontrollable}, VividAnimator adopts a unified diffusion framework based on ReferenceNet, conditioned on audio, hand poses, and a reference image.
We extract audio features using a pretrained Wav2Vec2~\cite{wav2vec2} model and process them with a \textbf{Dual-Stream Audio-Aware (DSAA)} module to enhance motion representation. In parallel, hand pose sequences are obtained via DWPose~\cite{yang2023effective} and encoded through a dedicated \textbf{Hand Encoder}, whose output is concatenated with the noise latent and injected into the denoising U-Net.
The reference image serves two purposes. First, it is encoded into a latent vector via VAE and passed through a U-Net-style \textbf{ReferenceNet} to extract identity features, which are injected via cross attention to ensure identity consistency. Second, the cropped hands are processed by the \textbf{Hand Clarity Codebook (HCC)}, which produces quantized discrete embeddings that are integrated into the denoising process through an additional attention stream.
To model temporal consistency, we incorporate a pretrained \textbf{motion module}~\cite{guo2023animatediff} that operates on the latent sequence to capture frame-to-frame dynamics.

%-------------------------------------------------------
\subsubsection{Dual-Stream Audio-Aware Module}

In this section, we design a \textbf{Dual-Stream Audio-Aware(DSAA)} module. As illustrated in Fig.~\ref{fig:2}, DSAA decouples the modeling of audio-driven motion into two complementary pathways, enabling more effective and specialized processing of motion dynamics. The Head Aware Stream captures global rhythm and prosodic patterns that drive natural head dynamics, while the Local Lip-Sync Stream ensures fine-grained alignment between phonemes and lip movements. 
This dual-path design enables the model to simultaneously achieve macro-level rhythmic motion and micro-level temporal synchronization.

\paragraph{Head Aware Stream.}  

We begin by extracting audio features $\mathbf{W_a} \in \mathbb{R}^{B \times F \times T \times D}$ from a pretrained speech encoder, where $B$ denotes the batch size, $F=24$ is the temporal window size, $T=50$ is the number of audio tokens, and $D=384$ is the feature dimension. Although these features contain rich temporal cues, they also include frame-wise fluctuations that are not essential for modeling global motion rhythm. To mitigate this, we apply average pooling along the window dimension, yielding a temporally smoothed representation $\mathbf{F_r} \in \mathbb{R}^{B \times T \times D}$. The window size is set to 24 so that each audio window corresponds to approximately one second of input, which provides an effective granularity for capturing syllable-level prosody while suppressing noise.
A lightweight projection network $G_\theta(\cdot)$ is then applied to compress $\mathbf{F_r}$ into a compact representation, which is injected into the denoising U-Net through cross-attention layers localized to the head region. To constrain the spatial influence of audio, we introduce a binary attention mask $\mathbf{M}_{\text{head}}$ that selects latent tokens corresponding to the head area. The head-aware module reduces dimensionality and simultaneously learns a task-specific embedding space that enables effective cross-attention with visual features.

\paragraph{Local Lip-Sync Stream.}  
In contrast to the Head-Aware Stream, which injects temporally smoothed global rhythm features, the Local Lip-Sync Stream incorporates fine-grained audio information at the frame level. Specifically, we condition each U-Net block on audio tokens $\mathbf{F}_{\text{lip}} \in \mathbb{R}^{B \times T \times D}$, where $T$ denotes the number of audio tokens and $D$ the feature dimension. 
At each block, audio tokens $\mathbf{F}_{\text{lip}}$ are injected into the hidden states via a Cross-Attention mechanism. This process uses phoneme-level cues to condition the visual features, thereby ensuring accurate lip synchronization.
This design allows the two streams to complement each other: the global rhythm stream governs natural head dynamics, whereas the local stream enforces precise alignment between audio and lip motion.

%---------------------------------------------------

\subsubsection{Hand Clarity Codebook}  

While prior works such as CyberHost~\cite{lin2025cyberhost} and ShowMaker~\cite{yang2024showmaker} adopt online codebook training, we instead pretrain the hand codebook offline on a large-scale, high-quality dataset. This enables the model to capture stable and expressive hand representations without being affected by noisy supervision or entangled gradients from other regions. Offline training also reduces the risk of overfitting, as it decouples hand representation learning from the full generation pipeline.

To learn a discrete hand codebook, we adopt a VQ-VAE~\cite{vqvae} comprising an encoder, a vector quantizer, and a decoder. The encoder consists of residual and self-attention blocks and gradually downsamples the input into a 256-dimensional latent representation. A vector quantizer with nearest-neighbor assignment maps the continuous features to a discrete codebook of 1024 entries. The decoder symmetrically upsamples and reconstructs the image using residual blocks, enabling the model to capture high-frequency hand details while maintaining a compact and reusable discrete latent space. Additional training configurations, along with quantitative and qualitative comparisons, will be discussed in Section~\ref{sec:exp}.

\begin{figure}[H]
    \centering
    \includegraphics[width=\linewidth,trim=0cm 0cm 0cm 0cm, clip]{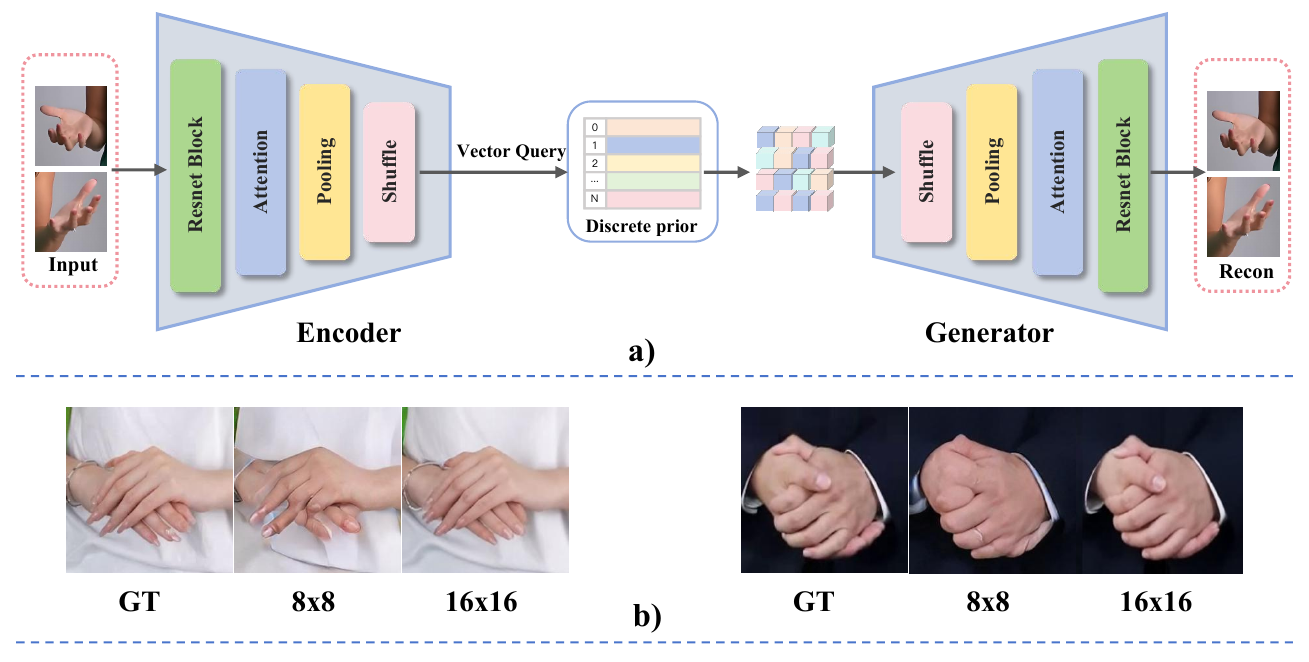}
    \caption{
    Overview of the Hand Clarity Codebook (HCC). a) The encoder discretizes cropped hand images into latent codes, and the generator reconstructs hands from these embeddings.
    b) Reconstructions with a 16×16 latent size preserve finer textures and structural details compared to 8×8.
    }
    \label{fig:3}
\end{figure}

\paragraph{Integration.}  
During denoising, we crop the left and right hand regions $I_l, I_r \in \mathbb{R}^{B \times C \times H \times W}$ from the reference image as inputs, where $B$ denotes the batch size, $C$ the number of channels, and $H \times W$ represents the spatial resolution of the cropped region. These images are encoded and quantized by the pretrained HCC, producing discrete hand embeddings:
\begin{equation}
[\mathbf{h}_q^l, \mathbf{h}_q^r] = \text{HCC}(I_l, I_r), 
\quad \mathbf{h}_q^l, \mathbf{h}_q^r \in \mathbb{R}^{B \times Q_{\text{dim}} \times Q_h \times Q_w},
\end{equation}
where $Q_{\text{dim}}$ denotes the codebook embedding dimension and $(Q_h, Q_w)$ represents the spatial resolution of the quantized grid. The left and right embeddings are then concatenated along the channel axis and flattened into a sequence of token embeddings $\mathbf{F_h} \in \mathbb{R}^{B \times (2 Q_h Q_w) \times Q_{\text{dim}}}$.

We inject these priors into the denoising U-Net via Codebook Attention, guiding the model to generate anatomically consistent and visually sharp hands. Given a hidden state $\tilde{\mathbf{h}}^{(l)} \in \mathbb{R}^{B \times N_l \times C_l}$ fused with audio features, where $N_l$ denotes the number of spatial positions and $C_l$ the channel dimension, the update rule is
\begin{equation}
\tilde{\mathbf{h}}^{(l)} \leftarrow 
\tilde{\mathbf{h}}^{(l)} +
\text{Attn}\!\left(\tilde{\mathbf{h}}^{(l)},\, \mathbf{F_h},\, \mathbf{F_h};\, \mathbf{M}_{\text{hand}}\right),
\end{equation}
where $\mathbf{M}_{\text{hand}} \in \{0,1\}^{B \times N_l \times 1}$ is a binary spatial mask that restricts attention to hand regions. This mechanism enables the diffusion process to directly exploit discrete, high-resolution hand priors, producing anatomically consistent hands with sharper contours and more realistic textures.
\subsubsection{Pose Calibration Trick}

In pose-driven generation, raw keypoints extracted from driving videos often suffer from scale mismatch, proportion distortion, or misalignment with the reference subject, which leads to unnatural or misplaced hands. To address this, we present the \textbf{Pose Calibration Trick (PCT)}, a plug-and-play, training-free component that reduces the rigid dependency of generated hands on the raw driving keypoints.  

PCT aligns the driving pose to the reference coordinate system through a three-step procedure.
First, a global scale normalization aligns overall body size by rescaling each keypoint $\mathbf{p}$ with horizontal and vertical factors $(r_x, r_y)$ estimated from torso anchors.
Second, segment-wise proportion adjustment enforces local geometric consistency, where correction factors $\rho_{ij} = l_{ij}^{\text{ref}} / l_{ij}$ ensure limb lengths match the reference.
Finally, an anchor-based translation shifts hand-related keypoints according to the torso center difference between driving and reference skeletons.
This process requires no additional parameters and can be seamlessly integrated into pose-driven generation frameworks.

\section{Experiments}
\label{sec:exp}

%-------------------------------------------------------------------------
\subsection{Training Setup}

\begin{figure*}[htbp]
    \centering
    \includegraphics[width=0.9\linewidth,trim=0.4cm 1.0cm 0.2cm 0cm, clip]{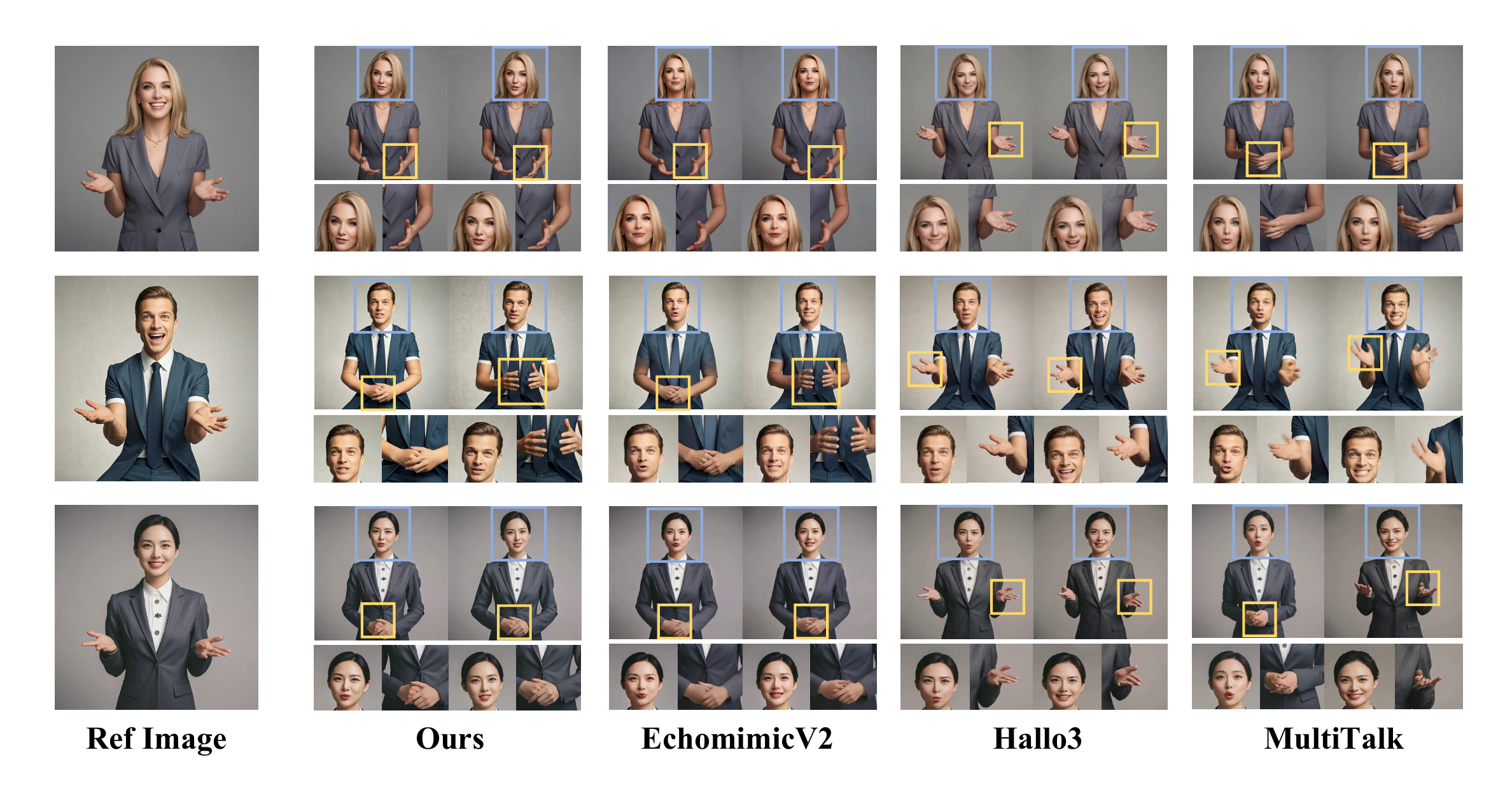}
    % \vspace{-1.0em}
    \caption{
    The results of VividAnimator compared with other audio-driven baselines.
    }
    \label{fig:6}
\end{figure*}

\paragraph{HCC Training Setup}

We pretrain the hand clarity codebook on a curated dataset of 260,000 high-quality hand images, each center cropped and resized to $256\times256$. All modules are trained for 200k iterations using the Adam optimizer with a fixed learning rate of $1\times10^{-4}$ and a batch size of 256. The training objective combines an $\ell_2$ reconstruction loss and the codebook commitment loss, with the latter weighted by $\beta = 0.25$. We adopt nearest-neighbor quantization with a codebook size of 1024 and an embedding dimension of 256, and experiment with both $8\times8$ and $16\times16$ latent grids.To assess the effect of codebook design, we compare offline pretraining with online joint training and evaluate different latent resolutions. 

\paragraph{Training Pipeline.}
Our training consists of two stages: single-frame training for reconstruction fidelity and multi-frame training for temporal consistency.
Both stages are optimized using latent loss.
During the multi-frame stage, the motion module is trained on a continuous sequence of 24 frames.This setup encourages temporal smoothness and cross-frame consistency in the generated results.

\paragraph{Implementation Details.}  
All experiments are conducted on 8 NVIDIA A800 GPUs.  
We set the batch size to 8 for single-frame training and 4 for multi-frame training.  
The learning rate is $1 \times 10^{-5}$ with AdamW optimizer. Both audio and image CFG scales are set to 2.5.

%--------------------------------------------------------

\begin{table*}[t]
  \centering
  \caption{
    Quantitative comparison with existing half-body animation methods.
  }
  \label{tab:quantitative} 
  \resizebox{\textwidth}{!}{
    \begin{tabular}{@{}c|ccccccccc@{}} 
      \toprule
      \text{Methods} &
      \text{SSIM$\uparrow$} &
      \text{PSNR$\uparrow$} &
      \text{CSIM$\uparrow$} &
      \text{FID$\downarrow$} &
      \text{FVD$\downarrow$} &
      \text{HKC$\uparrow$} &
      \text{HyperIQA$\uparrow$} &
      \text{Sync-C$\uparrow$} &
      \text{Sync-D$\downarrow$} \\ \midrule
      \text{Disco} &
      0.616 & 16.93 & 0.912 & 152.88 & 2311.18 & 0.784 & 57.60 & - & - \\
      \text{AnimateAnyone} &
      0.671 & 20.29 & 0.968 & 61.45 & 563.96 & 0.868 & 64.39 & 3.136 & 11.592 \\
      \text{MimicMotion} &
      0.689 & 20.01 & 0.961 & 91.39 & 855.11 & 0.910 & 59.60 & 5.047 & 9.843 \\
      \text{StableAnimator} &
      \textbf{0.733} & \textbf{21.29} & \textbf{0.974} & 60.92 & 334.12 & 0.896 & 61.23 & 5.184 & 9.433 \\
      \text{Hallo3} &
      0.679 & 18.64 & 0.933 & 106.01 & 642.54 & 0.861 & 55.82 & 2.687 & 12.226 \\
      \text{MultiTalk} &
      0.697 & 18.36 & 0.940 & 79.96 & 461.98 & 0.881 & 51.96 & 2.319 & 13.534 \\ \midrule 
      \text{EchomimicV2} &
      0.713 & 20.60 & 0.966 & 63.90 & 381.72 & 0.924 & 69.81 & 6.221 & 8.719 \\
      \text{VividAnimator} &
      0.711 & \underline{21.05} & \underline{0.970} &
      \textbf{54.43} & \textbf{333.45} & \textbf{0.942} &
      \textbf{71.04} & \textbf{6.241} & \textbf{8.446} \\ 
      \bottomrule
    \end{tabular}
  }
\end{table*}
\paragraph{Dataset}

We construct a corpus of about 200 hours of high-resolution videos from the internet. Among them, 40 hours consist of subtle-motion clips, and 160 hours contain large-motion clips, covering diverse speech and gesture patterns.  
To ensure data quality, we design a multi-stage filtering pipeline:  
(1) split raw videos into 5--10s clips, crop and center faces, and discard overly short samples;  
(2) remove clips with overlaid subtitles~\cite{yaofanguk_video_subtitle_remover};  
(3) eliminate videos with poor audio-visual synchronization~\cite{li2025latentsynctamingaudioconditionedlatent};  
(4) filter out severe motion-blurred clips;  
(5) apply super-resolution enhancement~\cite{rombach2022highresolutionimagesynthesislatent} to low-quality videos.  
Through this rigorous process, we obtain a dataset that is high-resolution, stable, and rich in motion diversity.
We further sample 130 clips to construct the test set, which will be publicly released to facilitate both quantitative and qualitative evaluation.  For inference, we employ FLUX~\cite{blackforest_flux} to synthesize 230 reference images, uniformly stratified by age, gender, and skin tone to ensure balanced coverage.

\begin{figure*}[t]
    \centering
    \includegraphics[width=\linewidth,trim=0.5cm 0.5cm 0.5cm 0cm, clip]{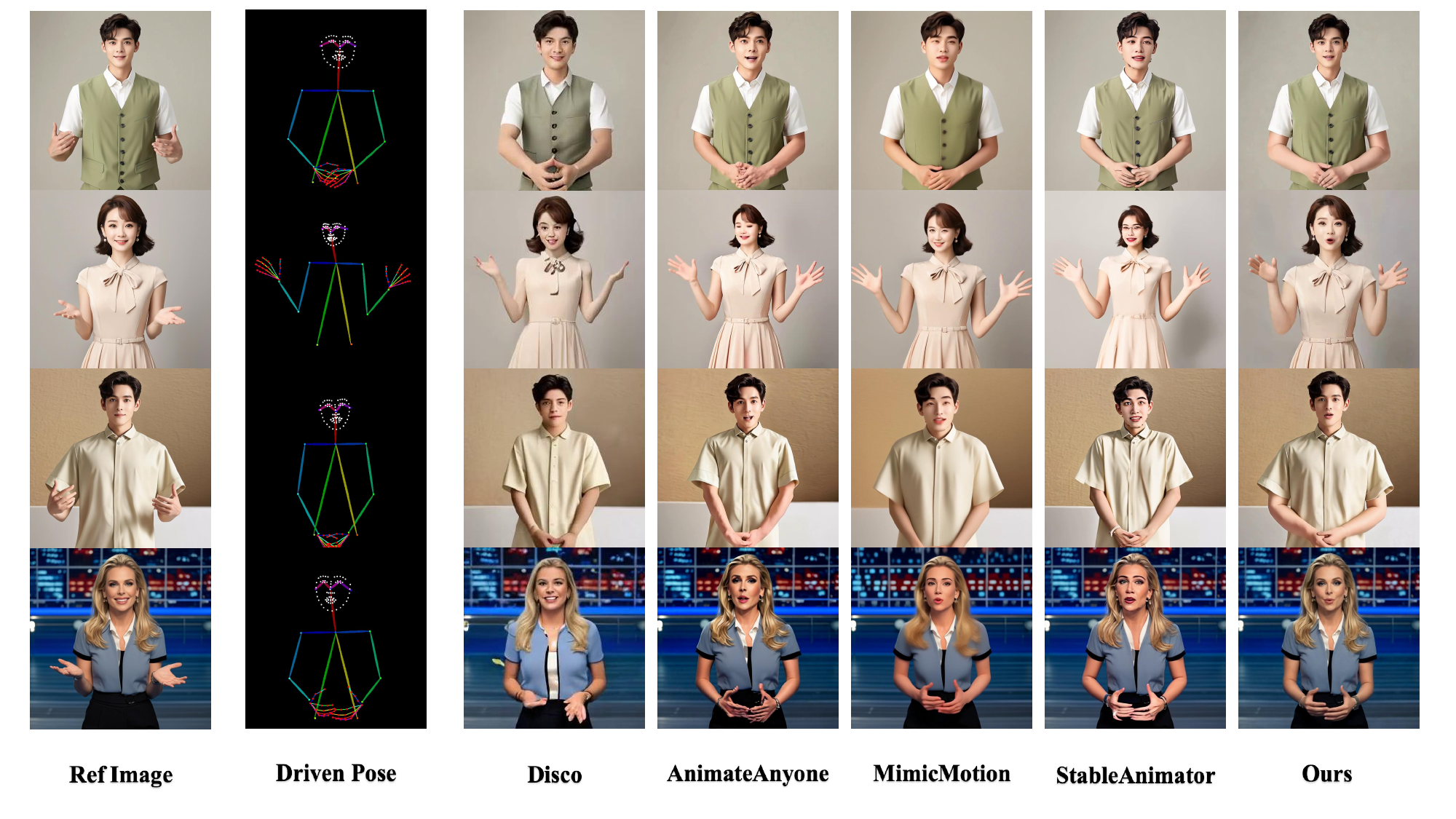}
    % \vspace{-1.0em}
    \caption{
    The results of VividAnimator compared with other pose-driven baselines.
    }
    \label{fig:4}
\end{figure*}

\subsection{Comparison with Existing Works}
\paragraph{Evaluation Metrics.}  
To comprehensively assess the effectiveness of our approach, we evaluate performance across four key aspects: image and video quality, identity consistency, motion clarity, and audio-visual synchronization.  
Specifically, image and video quality are measured using FID~\cite{fid}, PSNR~\cite{5596999}, SSIM~\cite{1284395}, FVD~\cite{unterthiner2019accurategenerativemodelsvideo} and  HyperIQA~\cite{Su_2020_CVPR} scores.  
Identity consistency is assessed by CSIM~\cite{csim}, which measures the semantic similarity between generated results and the reference image.  
For motion quality, we adopt HKC~\cite{lin2025cyberhost} (Hand Keypoints Confidence) and HKV~\cite{lin2025cyberhost} (Hand Keypoints Variance) to quantify the clarity and diversity of hand motions, and we define HMV (Head Motion Variance) to evaluate the variability of head dynamics.  
For audio-lip synchronization, we employ SyncNet~\cite{Prajwal_2020} to calculate Sync-C and Sync-D .  
Together, this evaluation suite covers static quality, dynamic coherence, and cross-modal alignment, providing a holistic reflection of realism, sharpness, and synchronization performance.  
\paragraph{Quantitative Results.}
We compare our method with representative baselines, including
Disco~\cite{wang2024discodisentangledcontrolrealistic},
Animate-Anyone~\cite{li2025latentsynctamingaudioconditionedlatent},
MimicMotion~\cite{zhang2025mimicmotionhighqualityhumanmotion},
StableAnimator~\cite{tu2024stableanimatorhighqualityidentitypreservinghuman},
EchoMimicV2~\cite{meng2025echomimicv2strikingsimplifiedsemibody},
Hallo3~\cite{cui2025hallo3highlydynamicrealistic},
and MultiTalk~\cite{kong2025lettalkaudiodrivenmultiperson}.
As reported in Tab.~\ref{tab:quantitative}, our approach achieves the best performance on perceptual and motion-oriented metrics, attaining the lowest FID and the highest HyperIQA and HKC. These results substantiate its superior visual quality and the ability to generate clearer and more stable hand gestures. 
Relative to EchoMimicV2, our model provides additional gains in HKC and HyperIQA, demonstrating the effectiveness of explicitly modeling hand motion. 
Although full-body pose–based methods (e.g., StableAnimator) yield marginally higher SSIM and PSNR, this outcome is expected since pixel-level metrics are biased toward dense pose supervision. In contrast, our method consistently outperforms all competing approaches on perceptual and motion-centric metrics, which more reliably capture the fidelity of speech-driven generation, thereby establishing its overall superiority.
\paragraph{Qualitative Results.}  
Beyond quantitative metrics, visual comparisons in Fig.~\ref{fig:4} and Fig.~\ref{fig:6} further highlight the effectiveness of our approach. Our method produces hand regions with sharper contours and richer structural details, while mitigating distortions and artifacts that commonly appear in high-frequency areas such as fingers. In addition, the overall perceptual quality is substantially enhanced: the synthesized videos look more natural and temporally stable, with improved preservation of reference identity. As shown in Fig.~\ref{fig:6}, unlike competing methods that often yield nearly static head poses or limited motion, our model generates clearer and more coherent head dynamics. 

\begin{figure}[H]
    \centering
    \includegraphics[width=\linewidth,trim=0.2cm 0.5cm 0.2cm 0cm, clip]{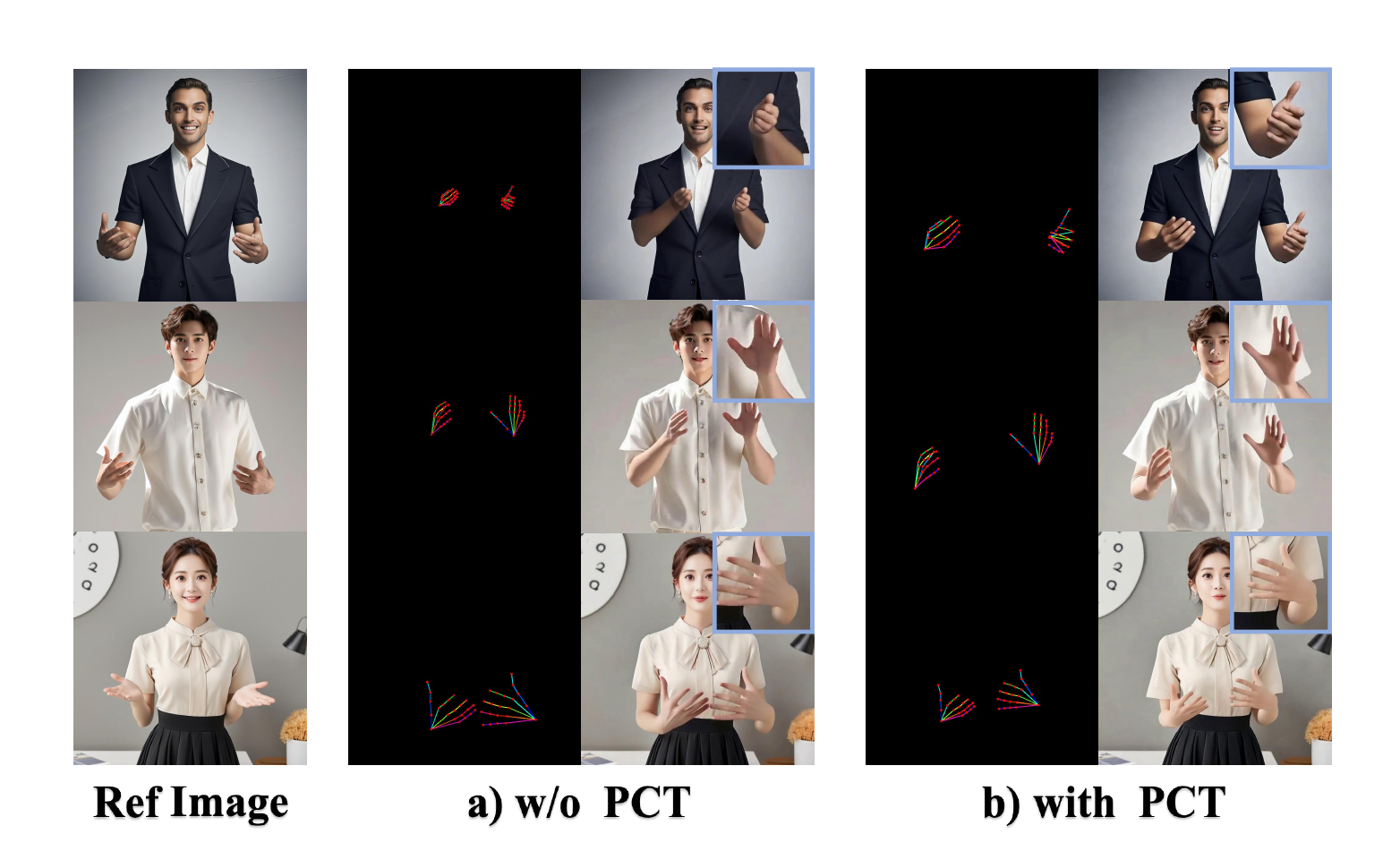}
    % \vspace{-1.0em}
    \caption{
    Effect of the Pose Calibration Trick (PCT).
    Given the same reference images and driving poses, VividAnimator with PCT b) produces hands that are better aligned in position and scale than the model without PCT a).
    }
    \label{fig:5}
\end{figure}

\subsection{Ablation Studies}

We further conduct ablation studies to validate the effectiveness of thress proposed components.  

\begin{table}[t]
\centering
\caption{Ablation study on VividAnimator across different metrics.}
\label{tab:ablation}
\resizebox{\columnwidth}{!}{ % 自动缩放到单栏宽度
\begin{tabular}{l|ccccccccccc}
\toprule
\text{Methods} & \text{HKC$\uparrow$} & \text{HKV$\uparrow$} & \text{HMV$\uparrow$} & \text{HyperIQA$\uparrow$}  \\
\midrule

HCC(online 8x8)    & 0.911 & 44.657 & 4.063 & 63.04 \\
HCC(online 16x16)  & 0.914 & 44.435 & 4.157 & 63.37 \\
HCC(offline 8x8)     & 0.920 & 44.495 & 4.095 & 65.24 \\
\midrule
w/o HCC(offline 16x16) & 0.907 & 43.749 & 3.636 & 69.40 \\
w/o Head Aware     & 0.922 & 44.641 & 3.295 & 69.65 \\
\midrule
EchomimicV2        & 0.922 & 42.221 & 3.061 & 69.81 \\
VividAnimator      & \textbf{0.942} & \textbf{46.452} & \textbf{4.435} & \textbf{71.04} \\
\bottomrule
\end{tabular}
}
\end{table}

\paragraph{Analysis of Hand Clarity Codebook.}  
Removing the HCC module leads to visibly blurrier hand regions and a substantial drop in HKC, indicating that the codebook is essential for modeling fine-grained hand details. We further examine how different design choices affect HCC effectiveness, including latent resolution and training strategy. As shown in Fig.~\ref{fig:3} and Tab.~\ref{tab:ablation}, the offline pretrained codebook with a $16\times16$ size achieves the best performance in terms of perceptual quality and detail preservation. Notably, all offline variants outperform their online counterparts, underscoring the benefit of decoupling codebook learning from the full pipeline. We attribute this to two factors: offline training learns stable tokens from large scale, high quality hand data under a single, well defined objective. In contrast, joint training exhibits a distributional bias. Hand regions are often small, blurry, occluded, and repetitive within the same subject or scene, which encourages overfitting to scene specific hand templates rather than a generalizable hand prior.

\paragraph{Analysis of  Dual-Stream Audio Aware Module.}  
When only a single audio injection stream is retained, the generated head motion lacks rhythm and lip synchronization becomes inaccurate.
As reported in Tab.~\ref{tab:ablation},the HMV score decreases, demonstrating that the dual-stream design plays a decisive role in achieving robust audio–visual alignment. Moreover, it also enriches head motion dynamics, leading to more expressive and natural visual results.

\paragraph{Analysis of Pose Calibration Trick.}  
When the Pose Calibration Trick is removed, the generated hand positions and scales deviate from the reference subject, leading to unnatural spatial proportions.  
As illustrated in Fig.~\ref{fig:5}, this module is crucial for improving perceptual naturalness and maintaining coherent hand-body coordination.  

\section{Conclusion}
\label{sec:conclusion}
In this work, we presented Vivid Animator, a novel end-to-end framework designed to overcome persistent challenges in human animation. Our method achieves high-fidelity half-body animation by simultaneously improving rendering quality and generating expressive, natural head movements.
Our framework is built upon three core technical components. First, the Hand Clarity Codebook (HCC) enhances fine-grained hand texture details by injecting rich hand priors from a pre-trained VAE model. Second, our Dual-Stream Audio-Aware Module (DSAA) enables the generation of more natural and expressive head movements by decoupling audio processing for lip and head motion. Finally, the Pose Calibration Trick (PCT) improves motion coherence in pose-driven methods by relaxing rigid constraints.
Collectively, these innovations allow Vivid Animator to generate state-of-the-art results in both visual quality and motion expressiveness. We believe our framework and its components lay a strong foundation for future research in controllable and high-fidelity human video synthesis.
{
    \small
    \bibliographystyle{ieeenat_fullname}
    \newpage
    \bibliography{main}

\begin{thebibliography}{60}
\providecommand{\natexlab}[1]{#1}
\providecommand{\url}[1]{\texttt{#1}}
\expandafter\ifx\csname urlstyle\endcsname\relax
  \providecommand{\doi}[1]{doi: #1}\else
  \providecommand{\doi}{doi: \begingroup \urlstyle{rm}\Url}\fi

\bibitem[Baevski et~al.(2020)Baevski, Zhou, Mohamed, and Auli]{wav2vec2}
Alexei Baevski, Henry Zhou, Abdelrahman Mohamed, and Michael Auli.
\newblock wav2vec 2.0: A framework for self-supervised learning of speech representations, 2020.

\bibitem[Bogo et~al.(2016)Bogo, Kanazawa, Lassner, Gehler, Romero, and Black]{Bogo:ECCV:2016}
Federica Bogo, Angjoo Kanazawa, Christoph Lassner, Peter Gehler, Javier Romero, and Michael~J. Black.
\newblock Keep it {SMPL}: Automatic estimation of {3D} human pose and shape from a single image.
\newblock In \emph{Computer Vision -- ECCV 2016}. Springer International Publishing, 2016.

\bibitem[Cai et~al.(2024)Cai, Yin, Zeng, Wei, Sun, Wang, Pang, Mei, Zhang, Zhang, Loy, Yang, and Liu]{cai2024smplerxscalingexpressivehuman}
Zhongang Cai, Wanqi Yin, Ailing Zeng, Chen Wei, Qingping Sun, Yanjun Wang, Hui~En Pang, Haiyi Mei, Mingyuan Zhang, Lei Zhang, Chen~Change Loy, Lei Yang, and Ziwei Liu.
\newblock Smpler-x: Scaling up expressive human pose and shape estimation, 2024.

\bibitem[Chang et~al.(2024)Chang, Shi, Gao, Fu, Xu, Song, Yan, Zhu, Yang, and Soleymani]{chang2024magicposerealistichumanposes}
Di Chang, Yichun Shi, Quankai Gao, Jessica Fu, Hongyi Xu, Guoxian Song, Qing Yan, Yizhe Zhu, Xiao Yang, and Mohammad Soleymani.
\newblock Magicpose: Realistic human poses and facial expressions retargeting with identity-aware diffusion, 2024.

\bibitem[Chen et~al.(2024)Chen, Cao, Chen, Li, and Ma]{echomimic}
Zhiyuan Chen, Jiajiong Cao, Zhiquan Chen, Yuming Li, and Chenguang Ma.
\newblock Echomimic: Lifelike audio-driven portrait animations through editable landmark conditions, 2024.

\bibitem[Cui et~al.(2025)Cui, Li, Zhan, Shang, Cheng, Ma, Mu, Zhou, Wang, and Zhu]{cui2025hallo3highlydynamicrealistic}
Jiahao Cui, Hui Li, Yun Zhan, Hanlin Shang, Kaihui Cheng, Yuqi Ma, Shan Mu, Hang Zhou, Jingdong Wang, and Siyu Zhu.
\newblock Hallo3: Highly dynamic and realistic portrait image animation with video diffusion transformer, 2025.

\bibitem[Deng et~al.(2022)Deng, Guo, Yang, Xue, Kotsia, and Zafeiriou]{ARCFACE}
Jiankang Deng, Jia Guo, Jing Yang, Niannan Xue, Irene Kotsia, and Stefanos Zafeiriou.
\newblock Arcface: Additive angular margin loss for deep face recognition.
\newblock \emph{IEEE Transactions on Pattern Analysis and Machine Intelligence}, 44\penalty0 (10):\penalty0 5962–5979, 2022.

\bibitem[Gan et~al.(2025)Gan, Yang, Zhu, Xue, and Hoi]{gan2025omniavatarefficientaudiodrivenavatar}
Qijun Gan, Ruizi Yang, Jianke Zhu, Shaofei Xue, and Steven Hoi.
\newblock Omniavatar: Efficient audio-driven avatar video generation with adaptive body animation, 2025.

\bibitem[Guan et~al.(2024)Guan, Yang, Wang, Zhou, He, Xu, Feng, Ding, Wang, Xie, Zhao, and Liu]{guan2024talkactenhancetexturalawareness2d}
Jiazhi Guan, Quanwei Yang, Kaisiyuan Wang, Hang Zhou, Shengyi He, Zhiliang Xu, Haocheng Feng, Errui Ding, Jingdong Wang, Hongtao Xie, Youjian Zhao, and Ziwei Liu.
\newblock Talk-act: Enhance textural-awareness for 2d speaking avatar reenactment with diffusion model, 2024.

\bibitem[Guo et~al.(2023)Guo, Yang, Rao, Liang, Wang, Qiao, Agrawala, Lin, and Dai]{guo2023animatediff}
Yuwei Guo, Ceyuan Yang, Anyi Rao, Zhengyang Liang, Yaohui Wang, Yu Qiao, Maneesh Agrawala, Dahua Lin, and Bo Dai.
\newblock Animatediff: Animate your personalized text-to-image diffusion models without specific tuning, 2023.

\bibitem[Güler et~al.(2018)Güler, Neverova, and Kokkinos]{denseposedensehumanpose}
Rıza~Alp Güler, Natalia Neverova, and Iasonas Kokkinos.
\newblock Densepose: Dense human pose estimation in the wild, 2018.

\bibitem[Ha et~al.(2019)Ha, Kersner, Kim, Seo, and Kim]{csim}
Sungjoo Ha, Martin Kersner, Beomsu Kim, Seokjun Seo, and Dongyoung Kim.
\newblock Marionette: Few-shot face reenactment preserving identity of unseen targets, 2019.

\bibitem[Heusel et~al.(2018)Heusel, Ramsauer, Unterthiner, Nessler, and Hochreiter]{fid}
Martin Heusel, Hubert Ramsauer, Thomas Unterthiner, Bernhard Nessler, and Sepp Hochreiter.
\newblock Gans trained by a two time-scale update rule converge to a local nash equilibrium, 2018.

\bibitem[Horé and Ziou(2010)]{5596999}
Alain Horé and Djemel Ziou.
\newblock Image quality metrics: Psnr vs. ssim.
\newblock In \emph{2010 20th International Conference on Pattern Recognition}, pages 2366--2369, 2010.

\bibitem[Hu et~al.(2024)Hu, Gao, Zhang, Sun, Zhang, and Bo]{hu2024animateanyoneconsistentcontrollable}
Li Hu, Xin Gao, Peng Zhang, Ke Sun, Bang Zhang, and Liefeng Bo.
\newblock Animate anyone: Consistent and controllable image-to-video synthesis for character animation, 2024.

\bibitem[Jiang et~al.(2025)Jiang, Liang, Yang, Lin, Zhong, and Zheng]{loopy}
Jianwen Jiang, Chao Liang, Jiaqi Yang, Gaojie Lin, Tianyun Zhong, and Yanbo Zheng.
\newblock Loopy: Taming audio-driven portrait avatar with long-term motion dependency, 2025.

\bibitem[Kingma and Welling(2022)]{kingma2022autoencodingvariationalbayes}
Diederik~P Kingma and Max Welling.
\newblock Auto-encoding variational bayes, 2022.

\bibitem[Kong et~al.(2025)Kong, Gao, Zhang, Kang, Wei, Cai, Chen, and Luo]{kong2025lettalkaudiodrivenmultiperson}
Zhe Kong, Feng Gao, Yong Zhang, Zhuoliang Kang, Xiaoming Wei, Xunliang Cai, Guanying Chen, and Wenhan Luo.
\newblock Let them talk: Audio-driven multi-person conversational video generation, 2025.

\bibitem[Labs(2024)]{blackforest_flux}
Black~Forest Labs.
\newblock Flux: Official inference repository for flux.1 models.
\newblock GitHub repository, 2024.
\newblock [Software].

\bibitem[Li et~al.(2025)Li, Zhang, Xu, Lin, Xie, Feng, Peng, Chen, and Xing]{li2025latentsynctamingaudioconditionedlatent}
Chunyu Li, Chao Zhang, Weikai Xu, Jingyu Lin, Jinghui Xie, Weiguo Feng, Bingyue Peng, Cunjian Chen, and Weiwei Xing.
\newblock Latentsync: Taming audio-conditioned latent diffusion models for lip sync with syncnet supervision, 2025.

\bibitem[Li et~al.(2023)Li, Qin, Liang, and Wei]{hdtr}
Yongyuan Li, Xiuyuan Qin, Chao Liang, and Mingqiang Wei.
\newblock Hdtr-net: A real-time high-definition teeth restoration network for arbitrary talking face generation methods.
\newblock In \emph{Chinese Conference on Pattern Recognition and Computer Vision (PRCV)}, pages 89--103. Springer, 2023.

\bibitem[Liang et~al.(2024)Liang, Jiang, Zhong, Lin, Rong, Yang, and Zhu]{liang2024superior}
Chao Liang, Jianwen Jiang, Tianyun Zhong, Gaojie Lin, Zhengkun Rong, Jiaqi Yang, and Yongming Zhu.
\newblock Superior and pragmatic talking face generation with teacher-student framework.
\newblock \emph{arXiv preprint arXiv:2403.17883}, 2024.

\bibitem[Lin et~al.(2025{\natexlab{a}})Lin, Jiang, Liang, Zhong, Yang, Zheng, and Zheng]{lin2025cyberhost}
Gaojie Lin, Jianwen Jiang, Chao Liang, Tianyun Zhong, Jiaqi Yang, Zerong Zheng, and Yanbo Zheng.
\newblock Cyberhost: A one-stage diffusion framework for audio-driven talking body generation.
\newblock In \emph{The Thirteenth International Conference on Learning Representations}, 2025{\natexlab{a}}.

\bibitem[Lin et~al.(2025{\natexlab{b}})Lin, Jiang, Yang, Zheng, and Liang]{lin2025omnihuman1rethinkingscalinguponestage}
Gaojie Lin, Jianwen Jiang, Jiaqi Yang, Zerong Zheng, and Chao Liang.
\newblock Omnihuman-1: Rethinking the scaling-up of one-stage conditioned human animation models, 2025{\natexlab{b}}.

\bibitem[Lin et~al.(2021)Lin, Wang, and Liu]{lin2021meshgraphormer}
Kevin Lin, Lijuan Wang, and Zicheng Liu.
\newblock Mesh graphormer, 2021.

\bibitem[Liu et~al.(2024)Liu, Yang, Akiyama, Huang, Li, Kuriyama, and Taketomi]{liu2024tangocospeechgesturevideo}
Haiyang Liu, Xingchao Yang, Tomoya Akiyama, Yuantian Huang, Qiaoge Li, Shigeru Kuriyama, and Takafumi Taketomi.
\newblock Tango: Co-speech gesture video reenactment with hierarchical audio motion embedding and diffusion interpolation, 2024.

\bibitem[Lu et~al.(2024)Lu, Xu, Zhang, Wang, and Tao]{lu2024handrefinerrefiningmalformedhands}
Wenquan Lu, Yufei Xu, Jing Zhang, Chaoyue Wang, and Dacheng Tao.
\newblock Handrefiner: Refining malformed hands in generated images by diffusion-based conditional inpainting, 2024.

\bibitem[Meng et~al.(2025{\natexlab{a}})Meng, Wang, Wu, Zheng, Li, and Ma]{meng2025echomimicv3}
Rang Meng, Yan Wang, Weipeng Wu, Ruobing Zheng, Yuming Li, and Chenguang Ma.
\newblock Echomimicv3: 1.3b parameters are all you need for unified multi-modal and multi-task human animation, 2025{\natexlab{a}}.

\bibitem[Meng et~al.(2025{\natexlab{b}})Meng, Zhang, Li, and Ma]{meng2025echomimicv2strikingsimplifiedsemibody}
Rang Meng, Xingyu Zhang, Yuming Li, and Chenguang Ma.
\newblock Echomimicv2: Towards striking, simplified, and semi-body human animation, 2025{\natexlab{b}}.

\bibitem[Pavlakos et~al.(2019)Pavlakos, Choutas, Ghorbani, Bolkart, Osman, Tzionas, and Black]{pavlakos2019expressivebodycapture3d}
Georgios Pavlakos, Vasileios Choutas, Nima Ghorbani, Timo Bolkart, Ahmed A.~A. Osman, Dimitrios Tzionas, and Michael~J. Black.
\newblock Expressive body capture: 3d hands, face, and body from a single image, 2019.

\bibitem[Pavlakos et~al.(2023)Pavlakos, Shan, Radosavovic, Kanazawa, Fouhey, and Malik]{hamer}
Georgios Pavlakos, Dandan Shan, Ilija Radosavovic, Angjoo Kanazawa, David Fouhey, and Jitendra Malik.
\newblock Reconstructing hands in 3d with transformers, 2023.

\bibitem[Prajwal et~al.(2020)Prajwal, Mukhopadhyay, Namboodiri, and Jawahar]{Prajwal_2020}
K~R Prajwal, Rudrabha Mukhopadhyay, Vinay~P. Namboodiri, and C.V. Jawahar.
\newblock A lip sync expert is all you need for speech to lip generation in the wild.
\newblock In \emph{Proceedings of the 28th ACM International Conference on Multimedia}, page 484–492. ACM, 2020.

\bibitem[Qi et~al.(2025)Qi, Ji, Xu, Zhang, Zhang, and Bo]{qi2025chatanyonestylizedrealtimeportrait}
Jinwei Qi, Chaonan Ji, Sheng Xu, Peng Zhang, Bang Zhang, and Liefeng Bo.
\newblock Chatanyone: Stylized real-time portrait video generation with hierarchical motion diffusion model, 2025.

\bibitem[Rombach et~al.(2022)Rombach, Blattmann, Lorenz, Esser, and Ommer]{rombach2022highresolutionimagesynthesislatent}
Robin Rombach, Andreas Blattmann, Dominik Lorenz, Patrick Esser, and Björn Ommer.
\newblock High-resolution image synthesis with latent diffusion models, 2022.

\bibitem[Shao et~al.(2024)Shao, Pang, Zheng, Sun, and Liu]{shao2024human4dit}
Ruizhi Shao, Youxin Pang, Zerong Zheng, Jingxiang Sun, and Yebin Liu.
\newblock Human4dit: 360-degree human video generation with 4d diffusion transformer.
\newblock \emph{ACM Transactions on Graphics (TOG)}, 43\penalty0 (6), 2024.

\bibitem[Su et~al.(2020)Su, Yan, Zhu, Zhang, Ge, Sun, and Zhang]{Su_2020_CVPR}
Shaolin Su, Qingsen Yan, Yu Zhu, Cheng Zhang, Xin Ge, Jinqiu Sun, and Yanning Zhang.
\newblock Blindly assess image quality in the wild guided by a self-adaptive hyper network.
\newblock In \emph{Proceedings of the IEEE/CVF Conference on Computer Vision and Pattern Recognition (CVPR)}, 2020.

\bibitem[Tian et~al.(2025)Tian, Hu, Wang, Zhang, and Bo]{tian2025emo2endeffectorguidedaudiodriven}
Linrui Tian, Siqi Hu, Qi Wang, Bang Zhang, and Liefeng Bo.
\newblock Emo2: End-effector guided audio-driven avatar video generation, 2025.

\bibitem[Tu et~al.(2024)Tu, Xing, Han, Cheng, Dai, Luo, and Wu]{tu2024stableanimatorhighqualityidentitypreservinghuman}
Shuyuan Tu, Zhen Xing, Xintong Han, Zhi-Qi Cheng, Qi Dai, Chong Luo, and Zuxuan Wu.
\newblock Stableanimator: High-quality identity-preserving human image animation, 2024.

\bibitem[Unterthiner et~al.(2019)Unterthiner, van Steenkiste, Kurach, Marinier, Michalski, and Gelly]{unterthiner2019accurategenerativemodelsvideo}
Thomas Unterthiner, Sjoerd van Steenkiste, Karol Kurach, Raphael Marinier, Marcin Michalski, and Sylvain Gelly.
\newblock Towards accurate generative models of video: A new metric \& challenges, 2019.

\bibitem[van~den Oord et~al.(2018)van~den Oord, Vinyals, and Kavukcuoglu]{vqvae}
Aaron van~den Oord, Oriol Vinyals, and Koray Kavukcuoglu.
\newblock Neural discrete representation learning, 2018.

\bibitem[Wang et~al.(2025)Wang, Wang, Jiang, and Xu]{wang2025fantasytalking2timesteplayeradaptivepreference}
MengChao Wang, Qiang Wang, Fan Jiang, and Mu Xu.
\newblock Fantasytalking2: Timestep-layer adaptive preference optimization for audio-driven portrait animation, 2025.

\bibitem[Wang et~al.(2024{\natexlab{a}})Wang, Li, Lin, Zhai, Lin, Yang, Zhang, Liu, and Wang]{wang2024discodisentangledcontrolrealistic}
Tan Wang, Linjie Li, Kevin Lin, Yuanhao Zhai, Chung-Ching Lin, Zhengyuan Yang, Hanwang Zhang, Zicheng Liu, and Lijuan Wang.
\newblock Disco: Disentangled control for realistic human dance generation, 2024{\natexlab{a}}.

\bibitem[Wang et~al.(2024{\natexlab{b}})Wang, Zhang, Gao, Wang, Zhou, Zhang, Yan, and Sang]{wang2024unianimate}
Xiang Wang, Shiwei Zhang, Changxin Gao, Jiayu Wang, Xiaoqiang Zhou, Yingya Zhang, Luxin Yan, and Nong Sang.
\newblock Unianimate: Taming unified video diffusion models for consistent human image animation.
\newblock \emph{arXiv preprint arXiv:2406.01188}, 2024{\natexlab{b}}.

\bibitem[Wang et~al.(2004)Wang, Bovik, Sheikh, and Simoncelli]{1284395}
Zhou Wang, A.C. Bovik, H.R. Sheikh, and E.P. Simoncelli.
\newblock Image quality assessment: from error visibility to structural similarity.
\newblock \emph{IEEE Transactions on Image Processing}, 13\penalty0 (4):\penalty0 600--612, 2004.

\bibitem[Wei et~al.(2024)Wei, Yang, and Wang]{wei2024aniportrait}
Huawei Wei, Zejun Yang, and Zhisheng Wang.
\newblock Aniportrait: Audio-driven synthesis of photorealistic portrait animations, 2024.

\bibitem[Xiang et~al.(2024)Xiang, Guo, Hu, Guo, Yuan, and Zhang]{shottalkwholebodytalking}
Jun Xiang, Yudong Guo, Leipeng Hu, Boyang Guo, Yancheng Yuan, and Juyong Zhang.
\newblock One shot, one talk: Whole-body talking avatar from a single image, 2024.

\bibitem[Xu et~al.(2024)Xu, Li, Su, Shang, Zhang, Liu, Wang, Yao, and Zhu]{hallo}
Mingwang Xu, Hui Li, Qingkun Su, Hanlin Shang, Liwei Zhang, Ce Liu, Jingdong Wang, Yao Yao, and Siyu Zhu.
\newblock Hallo: Hierarchical audio-driven visual synthesis for portrait image animation, 2024.

\bibitem[Xu et~al.(2023)Xu, Zhang, Liew, Yan, Liu, Zhang, Feng, and Shou]{xu2023magicanimatetemporallyconsistenthuman}
Zhongcong Xu, Jianfeng Zhang, Jun~Hao Liew, Hanshu Yan, Jia-Wei Liu, Chenxu Zhang, Jiashi Feng, and Mike~Zheng Shou.
\newblock Magicanimate: Temporally consistent human image animation using diffusion model, 2023.

\bibitem[Yang et~al.(2024{\natexlab{a}})Yang, Guan, Wang, Yu, Chu, Zhou, Feng, Feng, Ding, Wang, and Xie.]{yang2024showmaker}
Quanwei Yang, Jiazhi Guan, Kaisiyuan Wang, Lingyun Yu, Wenqing Chu, Hang Zhou, Zhiqiang Feng, Haocheng Feng, Errui Ding, Jingdong Wang, and Hongtao Xie.
\newblock Showmaker: Creating high-fidelity 2d human video via fine-grained diffusion modeling.
\newblock \emph{NeurIPS}, 2024{\natexlab{a}}.

\bibitem[Yang et~al.(2024{\natexlab{b}})Yang, Li, Wu, Jing, Li, Ji, Liang, Fan, and Wang]{yang2024megactorsigmaunlockingflexiblemixedmodal}
Shurong Yang, Huadong Li, Juhao Wu, Minhao Jing, Linze Li, Renhe Ji, Jiajun Liang, Haoqiang Fan, and Jin Wang.
\newblock Megactor-$\sigma$: Unlocking flexible mixed-modal control in portrait animation with diffusion transformer, 2024{\natexlab{b}}.

\bibitem[Yang et~al.(2023)Yang, Zeng, Yuan, and Li]{yang2023effective}
Zhendong Yang, Ailing Zeng, Chun Yuan, and Yu Li.
\newblock Effective whole-body pose estimation with two-stages distillation.
\newblock \emph{arXiv preprint arXiv:2307.15880}, 2023.

\bibitem[Yao et~al.(2025)Yao, Zhang, Xia, Wang, Roy-Chowdhury, and Li]{sonic}
Jianpeng Yao, Xiaopan Zhang, Yu Xia, Zejin Wang, Amit~K. Roy-Chowdhury, and Jiachen Li.
\newblock Sonic: Safe social navigation with adaptive conformal inference and constrained reinforcement learning, 2025.

\bibitem[YaoFANGUK(2023)]{yaofanguk_video_subtitle_remover}
YaoFANGUK.
\newblock video-subtitle-remover.
\newblock GitHub repository, 2023.
\newblock [Software].

\bibitem[Yu et~al.(2024)Yu, Huang, Cheng, and Birdal]{yu2024signavatarslargescale3dsign}
Zhengdi Yu, Shaoli Huang, Yongkang Cheng, and Tolga Birdal.
\newblock Signavatars: A large-scale 3d sign language holistic motion dataset and benchmark, 2024.

\bibitem[Zhang et~al.(2023)Zhang, Li, Li, Liu, Xue, Zhang, Jiang, Huang, Wang, and Wang]{emo}
Jiangning Zhang, Xiangtai Li, Jian Li, Liang Liu, Zhucun Xue, Boshen Zhang, Zhengkai Jiang, Tianxin Huang, Yabiao Wang, and Chengjie Wang.
\newblock Rethinking mobile block for efficient attention-based models.
\newblock In \emph{Proceedings of the IEEE/CVF International Conference on Computer Vision}, pages 1389--1400, 2023.

\bibitem[Zhang and Agrawala(2023)]{zhang2023adding}
Lvmin Zhang and Maneesh Agrawala.
\newblock Adding conditional control to text-to-image diffusion models, 2023.

\bibitem[Zhang et~al.(2025)Zhang, Gu, Wang, Wang, Cheng, Zhu, and Zou]{zhang2025mimicmotionhighqualityhumanmotion}
Yuang Zhang, Jiaxi Gu, Li-Wen Wang, Han Wang, Junqi Cheng, Yuefeng Zhu, and Fangyuan Zou.
\newblock Mimicmotion: High-quality human motion video generation with confidence-aware pose guidance, 2025.

\bibitem[Zhou et~al.(2024)Zhou, Wang, Chen, Bai, Li, Zhang, Xu, Yang, and Wang]{zhou2024realisdanceequipcontrollablecharacter}
Jingkai Zhou, Benzhi Wang, Weihua Chen, Jingqi Bai, Dongyang Li, Aixi Zhang, Hao Xu, Mingyang Yang, and Fan Wang.
\newblock Realisdance: Equip controllable character animation with realistic hands, 2024.

\bibitem[Zhu et~al.(2024)Zhu, Chen, Dai, Su, Xu, Cao, Yao, Zhu, and Zhu]{zhu2024champcontrollableconsistenthuman}
Shenhao Zhu, Junming~Leo Chen, Zuozhuo Dai, Qingkun Su, Yinghui Xu, Xun Cao, Yao Yao, Hao Zhu, and Siyu Zhu.
\newblock Champ: Controllable and consistent human image animation with 3d parametric guidance, 2024.

\bibitem[Zhuang et~al.(2024)Zhuang, Li, Chen, Wang, Liu, Qiao, and Wang]{zhuang2024vlogger}
Shaobin Zhuang, Kunchang Li, Xinyuan Chen, Yaohui Wang, Ziwei Liu, Yu Qiao, and Yali Wang.
\newblock Vlogger: Make your dream a vlog.
\newblock \emph{arXiv preprint arXiv:2401.09414}, 2024.

\end{thebibliography}
}

\end{document}